\documentclass[11pt,a4paper]{article}
\usepackage[hyperref]{emnlp2020}
\usepackage{times}
\usepackage{latexsym}

\usepackage{microtype}

\aclfinalcopy 
\def\aclpaperid{1709} 

\usepackage{bm}
\usepackage{amsmath}
\usepackage{array}
\usepackage{graphicx}
\usepackage{multirow}
\usepackage{makecell}
\usepackage{stfloats}
\usepackage{color}
\usepackage{CJKutf8}

\newcommand{\PreserveBackslash}[1]{\let\temp=\\#1\let\\=\temp}
\newcolumntype{C}[1]{>{\PreserveBackslash\centering}p{#1}}
\newcolumntype{R}[1]{>{\PreserveBackslash\raggedleft}p{#1}}
\newcolumntype{L}[1]{>{\PreserveBackslash\raggedright}p{#1}}

\newcommand{\chinese}[1]{\begin{CJK}{UTF8}{gkai}{}#1\end{CJK}}

\title{Toward Cross-Lingual Definition Generation for Language Learners}

\author{
	Cunliang Kong\textsuperscript{\dag}, 
	Liner Yang\textsuperscript{\dag},
	Tianzuo Zhang\textsuperscript{\dag},
	Qinan Fan\textsuperscript{\dag}, \\
	\textbf{Zhenghao Liu\textsuperscript{\ddag},
	Yun Chen\textsuperscript{\S},
	Erhong Yang\textsuperscript{\dag}} \\

	\textsuperscript{\dag}Beijing Language and Culture University, Beijing, China\\
	\textsuperscript{\ddag}Tsinghua University, Beijing, China\\
	\textsuperscript{\S}Shanghai University of Finance and Economics, Shanghai, China
}

\date{}

\begin{document}
\maketitle
\begin{abstract}
Generating dictionary definitions automatically can prove useful for language learners.
However, it's still a challenging task of cross-lingual definition generation.
In this work, we propose to generate definitions in English for words in various languages.
To achieve this, we present a simple yet effective approach based on publicly available pretrained language models.
In this approach, models can be directly applied to other languages after trained on the English dataset.
We demonstrate the effectiveness of this approach on zero-shot definition generation.
Experiments and manual analyses on newly constructed datasets show that our models have a strong cross-lingual transfer ability and can generate fluent English definitions for Chinese words.
We further measure the lexical complexity of generated and reference definitions.
The results show that the generated definitions are much simpler, which is more suitable for language learners.
\end{abstract}

\section{Introduction}

The definition modeling task proposed by \citet{Noraset2017DefinitionML} is to generate a dictionary definition of a specific word.
This task can prove useful for language learners, such as provide reading help by giving definitions for words in the text.
However, definition modeling can only work for a specific language, which puts high demands on users because it requires them to read definitions written in this language.
Besides, many low-resource languages lack large-scale dictionary data, making it difficult to train definition generation models for these languages.

\begin{figure}[t]
	\centering
	\includegraphics[width=\linewidth]{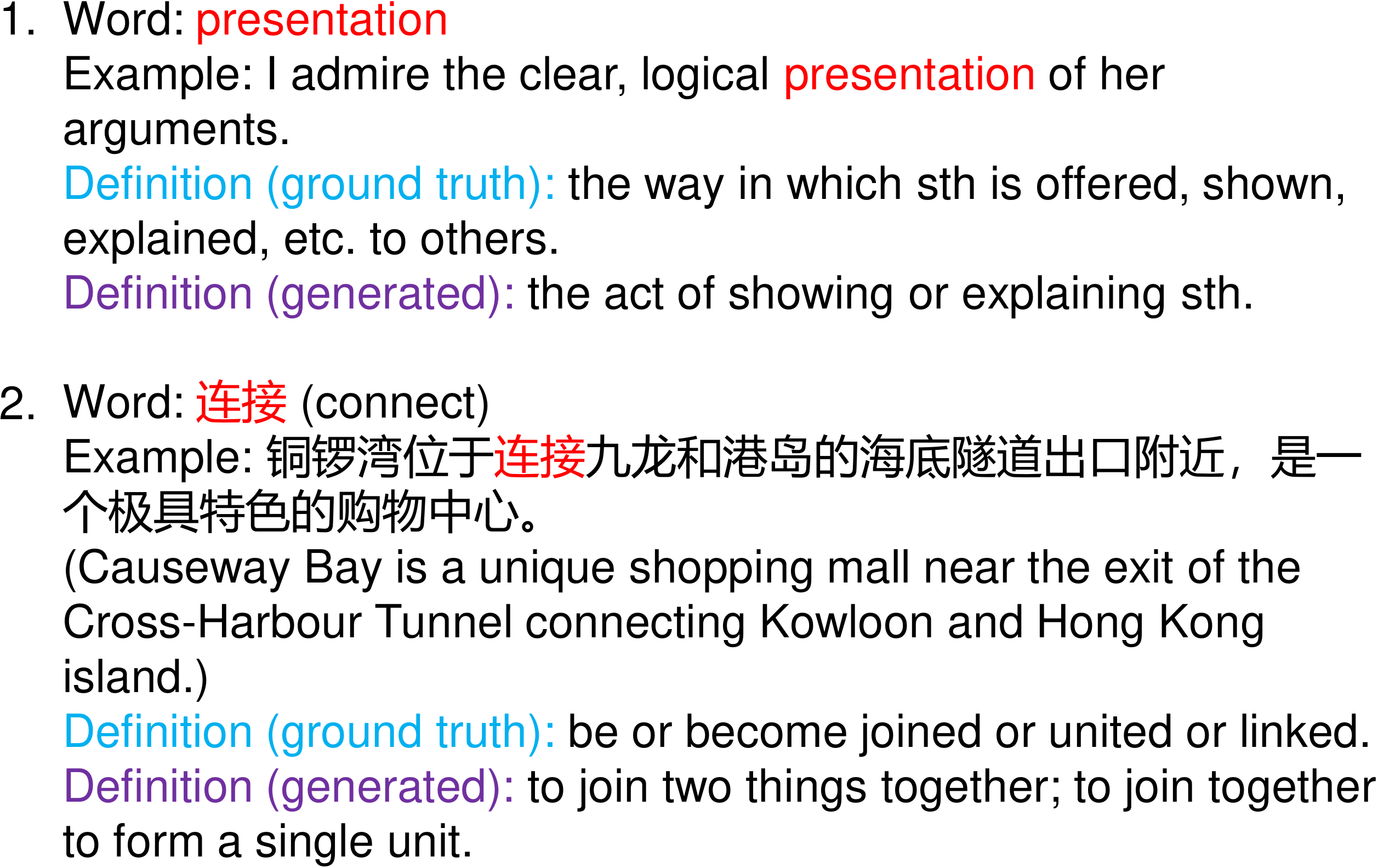}
	\caption{Generated definitions for English and Chinese words.}
	\label{fig:example}
\end{figure}

Therefore, we emphasize the necessity of generating definitions cross-lingually, which can generate definitions for various language inputs, as illustrated in figure \ref{fig:example}.
Since English is widely used around the world, and English dictionary resources are relatively easy to obtain, we choose to generate definitions in English.
In this way, a cross-lingual model trained on English can be directly applied to other languages.

The challenging issue is how to effectively transfer the knowledge of definition generation learned in English to other languages.
To solve this problem, we propose to employ cross-lingual pretrained language models \citep{Devlin2019BERTPO,Lample2019CrosslingualLM} as encoders.
These models have shown to be able to encode sequences of various languages, which enables the ability of cross-lingual transfer \citep{Chi2019CrossLingualNL,Ren2019ExplicitCP}.


To verify our proposed method, we build an English dataset for model training and a Chinese dataset for zero-shot cross-lingual evaluation.
Experiments and manual analyses on the constructed datasets show that our proposed models have good cross-lingual transfer ability.
Compared with the reference definitions in the CWN dataset, although the generated definitions are still insufficient on the accuracy, their fluency is already good enough.

Furthermore, considering the generated definitions are provided for language learners, and many of them are non-English native speakers, we argue that the difficulty of definitions should be under control.
We control the lexical complexity of generated definitions by limiting definitions in the training set to the \textit{Oxford 3000} vocabulary, which is a list of important and useful words that are carefully selected by language experts and experienced teachers \footnote{\url{https://www.oxfordlearnersdictionaries.com/about/oxford3000}}.
We compute four different metrics to measure the lexical complexity.
Definitions generated by our models outperform the reference definitions on all four metrics by a large margin.
The result shows that our method can generate simpler definitions, which is suitable for language learners.

\section{Method}
Our work follows the seq2seq approach but differs from previous work \citep{Noraset2017DefinitionML, Gadetsky2018ConditionalGO, Ishiwatari2019LearningTD, Yang2019IncorporatingSI} in adopting multilingual input sequences.
To facilitate a better understanding, we first introduce the basic framework for the definition modeling task. On this basis, we then describe our approaches for cross-lingual definition generation.

\subsection{Definition Modeling Task}
The goal of the definition modeling task is to generate definitions for given words.
Since words may have various meanings in different contexts, we also provide an example sentence containing the given word for the input following \citet{Gadetsky2018ConditionalGO} and \citet{Ishiwatari2019LearningTD}.

Formally, given a word $w$ and a corresponding example sentence $\bm{s}=[s_1, ..., s_m]$ containing the word, we employ a language model as follows to generate the definition word by word.
\begin{equation}
P(\bm{\hat{d}} |f(w, \bm{s})) = \prod_{i=1}^{n}P(\hat{d_i}|\hat{d_{<i}}, f(w, \bm{s})),
\end{equation}
where $\bm{\hat{d}} = [\hat{d1}, ..., \hat{d_n}]$ is the generated definitions, and $f(\cdot)$ is a nonlinear function employed to encode the given word and example sentence.

We train the parameters by minimizing the KL-divergence between the generated definitions and the true distributions underlying the dataset.


\subsection{Cross-Lingual Definition Generation}
Now we extend the definition modeling task to the cross-lingual scenario.
The only difference is that the input words and example sentences are multilingual.
For example, the model can not only generate English definitions for English words and example sentences ($<w^{en},\bm{s}^{en}> \rightarrow \hat{\bm{d}}^{en}$), but also generate English definitions for words and example sentences in other languages, such as Chinese ($<w^{zh},\bm{s}^{zh}> \rightarrow \hat{\bm{d}}^{en}$).

It has been shown that pretrained cross-lingual language models yield high quality cross-lingual representations and enable effective zero-shot transfer \citep{Lample2019CrosslingualLM,Ren2019ExplicitCP,Huang2019UnicoderAU,Sun2020UnsupervisedNM}.
Therefore, we alter the encoder $f(\cdot)$ as a pretrained cross-lingual language model to make it compatible with different languages.
In this way, the model can encode words and example sentences in any language into the same cross-lingual vector space.
We compare the effects of two language models in this work, namely multilingual BERT (mBERT) \citep{Devlin2019BERTPO} and XLM \citep{Lample2019CrosslingualLM}.

We concatenate words and example sentences ($[w; \bm{s}]$) as inputs for the encoder, with a special token $[SEP]$ as the separator.
In order to distinguish between tokens in words and example sentences, we set different token type ids.
The token type ids are 0 for given words, and 1 for example sentences.
Different type ids correspond to different embeddings, which can be obtained through training.
Token type embeddings are added to token embeddings before encoding, together with position embeddings.

We employ an affine layer on the output embeddings of the encoder and provide the results as memories to a transformer decoder \citep{Vaswani2017AttentionIA}, which is used to obtain definitions.

\section{Experiments}
In this section, we experimentally analyze the performance of our method. We train and fine-tune the entire model on the English dataset, and then apply the obtained model directly to the Chinese dataset. Our main question of interest is whether the proposed method is capable of generating definitions for untrained languages.

\begin{table}[ht]
	\setlength{\tabcolsep}{5pt}
	\centering
	\begin{tabular}{lrrR{1cm}R{1cm}}
		\hline
		Dataset & Words & Entries & Exp. & Def. \\
		\hline
		\hline
		OALD & & & & \\
		\hline
		Train & 17,146 & 69,738 & 6.81 & 12.44 \\
		Valid & 952 & 3,704 & 6.75 & 12.62 \\
		Test & 952 & 4,145 & 6.84 & 12.54 \\
		\hline
		\hline
		CWN & & & & \\
		\hline
		Test & 200 & 200 & 16.09 & 9.39 \\
		\hline
	\end{tabular}
	\caption{Statistics of the OALD dataset and the CWN dataset. The columns in the table are the number of words and entries, the average length of example sentences and definitions.}
	\label{table:data}
\end{table}

\subsection{Datasets}
We build two new datasets for this work, namely the OALD dataset and the CWN dataset. Note that definitions in both datasets are in English. We present the statistics of both datasets in table \ref{table:data}.

\paragraph{OALD} We collected data of the Oxford Advanced Learner's Dictionary \citep{Michael2010Oxford} from its official website \footnote{\url{https://www.oxfordlearnersdictionaries.com}}.
Each entry is a triplet containing the word, its definition and the example sentence of this word in the given meaning. The final dataset contains 19,050 words. Since each word may have multiple definitions, and each definition may correspond to multiple example sentences, the dataset contains 77,587 entries in total.

\paragraph{CWN} Chinese WordNet \footnote{\url{http://lope.linguistics.ntu.edu.tw/cwn2}} \citep{Huang2010Chinesewordnet} is a WordNet-like semantic lexicon, where each sense of word corresponds to a synset in WordNet. We extracted Chinese words, example sentences, and English definitions from it. The dataset contains 7,728 words and 68,497 entries. We randomly select 200 words to form a test set for this work.

\subsection{Setup}
The entire model is trained in two stages following \citet{Wang2019ToTO}. 
The main purpose of the first stage is to obtain a decoder with good performance.
Therefore, we fix the parameters of the pretrained language models, and only parameters in the decoder are learned.
In the second stage, we use a rather smaller learning rate to train all the parameters of the entire model.
This is to fine-tune the parameters, especially those in the pretrained language models, to obtain better results.
More details of the hyperparameters are in the Appendix \ref{apdx:params}.

We implement the proposed architecture based on the transformers library \citep{Wolf2019HuggingFacesTS}.
And we use fastText vectors \citep{Bojanowski2016enriching} to initialize embeddings of words in definitions.
We report the PPL and BLEU scores as previous work did \citep{Noraset2017DefinitionML, Gadetsky2018ConditionalGO, Ishiwatari2019LearningTD, Yang2019IncorporatingSI} on the OALD test set after each stage of training.

However, both scores can only measure literal similarity. On the CWN dataset, it's inappropriate to use these score to compare the generated Oxford-like definitions and the reference definitions from WordNet. Therefore, we organized manual evaluations to measure the effectiveness of models when given Chinese inputs.


\begin{table}[t]
	\setlength{\tabcolsep}{8pt}
	\centering
	\begin{tabular}{lrrr}
		\hline
		& PPL & BLEU & $\Delta$ BLEU \\
		\hline
		XLM-fix & 19.82 & 21.03 & \\
		mBERT-fix & 18.79 & 22.28 & \\
		XLM-ft & \textbf{16.37} & \textbf{23.10} & \textbf{+2.07} \\
		mBERT-ft & 18.28 & 22.56 & +0.28 \\
		\hline
	\end{tabular}
	\caption{PPL and BLEU scores on the OALD test set. Note that $\Delta$ BLEU is the improvements obtained after the fine-tune stage.}
	\label{table:oald_result}
\end{table}

\subsection{Monolingual Results}
Table \ref{table:oald_result} presents the performance of models on the OALD dataset.
The experiment results show that our proposed models have the ability to generate dictionary definitions for English words.

The mBERT model performs better than the XLM model without adjusting the pretrained parameters.
In this case, the PPL value of the mBERT model is 1.03 lower than that of the XLM model, and the BLEU score is 1.25 higher.
However, we also noticed the effect of fine-tuning on the performance of models.
By applying the fine-tune strategy, the PPL value of the XLM model is decreased by 2.42 and the BLEU score increased by 2.07, which exceeded the improvement obtained by the mBERT model. 
The fine-tuned XLM model is the best performing model on the OALD dataset.

\subsection{Zero-Shot Cross-Lingual Results}
For manual evaluation, we randomly shuffle a total of 800 definitions generated by 4 models and let 3 scorers rate them independently.
The English definitions in the dataset, which are extracted from the WordNet, are used as references when scoring.
Each scorer evaluates both the accuracy and fluency of each definition.
Accuracy evaluates whether the generated definitions explain the given words.
Fluency evaluates the quality of the definitions and whether they conform to grammar.
Both accuracy and fluency are scored from 1 to 3 points, with 1 being the lowest and 3 being the highest.
We calculate the average of these scores, and present them in table \ref{table:cwn_result}.
We also select some generated samples and put them in the Appendix \ref{apdx:sample}.

\begin{table}[t]
	\setlength{\tabcolsep}{5pt}
	\centering
	\begin{tabular}{l|lrrrr}
		\hline
		& & \#1 & \#2 & \#3 & Avg. \\
		\hline
		\multirow{4}{*}{Acc.} & XLM-fix & 1.10 & 1.11 & 1.06 & 1.09 \\
		& mBERT-fix & 1.69 & 1.63 & 1.30 & 1.54 \\
		& XLM-ft & 1.65 & 1.63 & 1.29 & 1.52 \\
		& mBERT-ft & \textbf{1.86} & \textbf{1.80} & \textbf{1.46} & \textbf{1.71} \\
		\hline
		\multirow{4}{*}{Flu.} & XLM-fix & 3.00 & 3.00 & 2.14 & 2.72 \\
		& mBERT-fix & 2.96 & 2.99 & 2.23 & 2.73 \\
		& XLM-ft & 2.96 & 2.98 & 2.22 & 2.72 \\
		& mBERT-ft & 2.96 & 2.97 & 2.25 & 2.73 \\
		\hline
	\end{tabular}
	\caption{Accuracy and fluency scores on the CWN dataset.}
	\label{table:cwn_result}
\end{table}

On the accuracy scores, the fine-tuned mBERT model outperforms all other models.
The results show that mBERT has stronger cross-lingual transfer ability than XLM.
Reinforcing the finding of \citet{Heijden2019ACO}, no additional benefit obtained in the Translation Language Modeling (TLM) method for zero-shot cross-lingual transfer.
On the fluency scores, it can be seen that all 4 models used in the experiments have high scores, for the encoders pretrained on large-scale corpora carries sufficient language information.
We don't think that the definitions generated by these models have distinctive differences on the fluency.

\begin{table}[t]
	\setlength{\tabcolsep}{7pt}
	\centering
	\begin{tabular}{l|rrrr}
		\hline
		& LD & LS & TTR & MSTTR \\
		\hline
		Reference & 0.52 & 0.36 & 0.41 & 0.78 \\
		mBERT-ft & \textbf{0.47} & \textbf{0.21} & \textbf{0.19} & \textbf{0.65} \\
		\hline
	\end{tabular}
	\caption{Lexical complexity measurements of generated and reference definitions.}
	\label{table:cpl}
\end{table}


Furthermore, we introduce 4 metrics for measuring the lexical complexity of definitions, namely lexical density (LD) \citep{halliday1985spoken}, lexical sophistication (LS) \citep{linnarud1986lexis}, type/token ratio (TTR) \citep{Richards1987TypeTokenRW}, and mean segmental TTR (MSTTR) \citep{Johnson1944Studies}.
Note that we use all 50-words segments to compute the MSTTR, and we employ a toolkit implemented by \citet{Lu2012TheRel} to compute these metrics.
As illustrated in table \ref{table:cpl}, the generated definitions outperform the reference on all of the 4 metrics, which indicate lower lexical complexity and suitable for language learners.

\section{Related Work}
This paper mainly related to two aspects of work, namely definition modeling and cross-lingual pretrained language models.

\citet{Noraset2017DefinitionML} first proposed the use of language models to generate definitions for given words.
Since their work can only generate one definition for one word, it cannot serve polysemies well.
\citet{Gadetsky2018ConditionalGO} introduced the context of words as input and computed the AdaGram vector \citep{Bartunov2016BreakingSA} for the given words to distinguish different meanings of it.
To make the model more interpretable, \citet{Chang2018xSenseLS} proposed to project the given words to high-dimensional sparse vectors, and picked different dimensions for different meanings.
While all previous work studied English, \citet{Yang2019IncorporatingSI} specifically explored definition modeling for Chinese words.
They incorporated \textit{sememes}, minimal semantic units, as part of the representation of given words to generate definitions.
\citet{Mickus2019MarkMW} implemented a transformer-based sequence-to-sequence model to train the definition generation model in an end-to-end fashion.
\citet{Ishiwatari2019LearningTD} extended this task to describe unknown phrases by using both local and global contexts.

\citet{Devlin2019BERTPO} released the multilingual BERT pretrained on corpora of 104 languages, which is capable of generating high quality cross-lingual representations. \citet{Lample2019CrosslingualLM} then introduced the TLM task into cross-lingual language model pretraining, and received SOTA results on cross-lingual classification and machine translation tasks. 
\citet{Wang2019CrossLingualBT} employed a linear transformation mechanism to generate cross-lingual contextualized word embeddings based on BERT models, and then used these embeddings for zero-shot dependency parsing.
\citet{Chi2019CrossLingualNL} proposed a novel pretrained model named XNLG for many-to-many corss-lingual NLG tasks.

\section{Conclusion}
In this work, we employ pretrained language models, namely mBERT and XLM for cross-lingual definition generation.
In addition, we propose to use the \textit{Oxford 3000} vocabulary to limit the lexical complexity of generated definitions.
We build the OALD dataset for monolingual training and the CWN dataset for cross-lingual evaluation.
Experiments indicate the strong cross-lingual transfer ability of our proposed method.
Furthermore, results on lexical complexity shows that definitions generated using our method is simpler than the reference, which is suitable for language learners.

\clearpage

\bibliographystyle{acl_natbib}
\bibliography{xdefgen}

\clearpage
\appendix
\begin{table*}[ht]
	\setlength{\tabcolsep}{10pt}
	\centering
	\begin{tabular}{L{2cm}|L{12cm}}
		\hline
		Input & \multirow{5}{12cm}{\chinese{具备} (possess) \\
			\chinese{其实创业者若未{\color{red}具备}相当的有利条件，常会落得乘兴而来、败兴而归的惨痛后果。 \\
			(In fact, if entrepreneurs don't possess considerable conditions, they will often have tragic consequences.)
		}} \\
		& \\
		& \\
		& \\
		& \\
		Reference & have or possess, either in a concrete or an abstract sense \\
		\hline
		XLM-fix & to make sth such as a vehicle or a vehicle move quickly and suddenly \\
		mBERT-fix & to have a particular effect or result \\
		XLM-ft & to have sth as a purpose or feature \\
		mBERT-ft & having a particular quality or feature \\
		\hline
		\hline
		Input & \multirow{4}{12cm}{\chinese{总数} (total number) \\
			\chinese{南非有800多种的鸟禽，为世人所知鸟类{\color{red}总数}的10\%。} \\
			(There are more than 800 bird species in South Africa, 10\% of the total number known to the world.)
		} \\
		& \\
		& \\
		& \\
		Reference & a quantity obtained by addition \\
		\hline
		XLM-fix & to make sth seem more important or better than it really is \\
		mBERT-fix & the number of things that a number or amount is divided into \\
		XLM-ft & the whole number of people or things \\
		mBERT-ft & the number of people or things that are counted in whole number \\
		\hline
		\hline
		Input & \multirow{4}{12cm}{\chinese{大幅} (significantly) \\
			\chinese{民家被指定为文化财产的案例{\color{red}大幅}增加，迄今已超过三百件。} \\
			(The number of cases where private homes have been designated as cultural property has increased significantly, with more than 300 cases to date.)
		} \\
		& \\
		& \\
		& \\
		Reference & to a great extent or degree \\
		\hline
		XLM-fix & to a greater degree than sth else \\
		mBERT-fix & more than usual or most important \\
		XLM-ft & to a large degree; in large amounts \\
		mBERT-ft & very large in size, amount, etc. \\
		\hline
	\end{tabular}
	\caption{Generated samples}
	\label{table:samples}
\end{table*}

\begin{table}[ht]
	\setlength{\tabcolsep}{10pt}
	\centering
	\begin{tabular}{l|r}
		\hline
		\textbf{Hyper-Parameter} & \textbf{Value} \\
		\hline
		decoder heads & 5 \\
		decoder layers & 6 \\
		decoder FFN layer units & 2,048 \\
		learning rate (1st stage) & 5e-4 \\
		warm-up (1st stage) & 4,000 \\
		learning rate (2nd stage) & 2e-5 \\
		warm-up (2nd stage) & 2,000 \\
		dropout & 0.2 \\
		clip norm & 0.1 \\
		batch size & 256 \\
		\hline
	\end{tabular}
	\caption{Hyperparameters used in Experiments}
	\label{table:params}
\end{table}

\section{Appendices}
\subsection{Hyperparameters in Experiments} \label{apdx:params}
As illustrated in table \ref{table:params}, the initial settings of all models in experiments are basically the same.
After obtaining the encoder output, we employ a transformer decoder with 5 heads, 6 layers to obtain the generated results.
The feedforward layer used in the decoder has 2,048 units.
In both training stages we set dropout as 0.2 and gradient norms are clipped to a maximum value of 0.1.

During training, we choose and save the model that performs best on the validation set. We consider the performance on two metrics of PPL and BLEU, and select the model with the highest BLEU after PPL no longer drops.

\subsection{Generated Samples} \label{apdx:sample}
Table \ref{table:samples} shows the generated samples in monolingual and zero-shot cross-lingual experiments.


\end{document}